\documentclass[11pt]{article}

\usepackage[final]{acl}
\usepackage{times}
\usepackage{latexsym}

\usepackage[T1]{fontenc}
\usepackage[utf8]{inputenc}

\usepackage{microtype}

\usepackage{inconsolata}

\usepackage{graphicx}

\usepackage{multirow}
\usepackage{pgf-pie}
\usepackage{url}
\usepackage{booktabs}

\title{ChakmaNMT: Machine Translation for a Low-Resource and Endangered Language via Transliteration}

\author{
 \textbf{Aunabil Chakma\textsuperscript{1}},
 \textbf{Aditya Chakma\textsuperscript{2}},
 \textbf{Masum Hasan\textsuperscript{3}},
 \textbf{Soham Khisa\textsuperscript{2}},
\\
 \textbf{Chumui Tripura\textsuperscript{4}},
 \textbf{Rifat Shahriyar\textsuperscript{2}}
\\
 \textsuperscript{1}University of Arizona,
 \textsuperscript{2}Bangladesh University of Engineering and Technology,
\\
 \textsuperscript{3}University of Rochester,
 \textsuperscript{4}Chittagong University of Engineering and Technology
\\
\texttt{aunabilchakma@arizona.edu, 1505120@ugrad.cse.buet.ac.bd,}\\
\texttt{1705120@ugrad.cse.buet.ac.bd, u1604131@student.cuet.ac.bd,}\\
\texttt{m.hasan@rochester.edu, rifat@cse.buet.ac.bd}
}
\begin{document}
% \newfontface\banglafont{SutonnyOMJ.ttf}
% \newfontface\chakmafont{ribenguni.ttf}
\maketitle
\begin{abstract}
We present the first systematic study of machine translation for Chakma, an endangered and extremely low-resource Indo-Aryan language, with the goal of supporting language access and preservation.
We introduce a new Chakma--Bangla parallel and monolingual dataset, along with a trilingual Chakma--Bangla--English benchmark for evaluation.
To address script mismatch and data scarcity, we propose a character-level transliteration framework that exploits the close orthographic and phonological relationship between Chakma and Bangla, preserving semantic content while enabling effective transfer from Bangla and multilingual pretrained models.
We benchmark from-scratch MT, fine-tuned pretrained models, and large language models via in-context learning.
Results show that transliteration is essential and that fine-tuning and in-context learning substantially outperform from-scratch baselines, with strong asymmetry across translation directions.
\end{abstract}

\section{Introduction}
The Chakma language is spoken by the indigenous Chakma people across Bangladesh, the easternmost regions of India, and western Myanmar, and belongs to the Indo-Aryan language family \cite{mohsin2013language}.
It is spoken by over 700,000 people across the region \cite{censusindia,archive}.
Despite this population, Chakma remains predominantly oral, with limited use of its writing system, leaving the language critically under-resourced in digital form.
As noted by \citet{Saikia}, Chakma is classified as “Definitely Endangered,” and continued language loss poses risks to cultural identity and community continuity.
Although revitalization efforts exist-such as pre-primary materials produced by the National Curriculum and Textbook Board of Bangladesh \cite{NCTB} and 
limited school textbooks in India \cite{CADC}-they remain largely confined to education, 
while everyday communication and public discourse increasingly rely on dominant regional languages such as Bangla, underscoring the lack of computational support for Chakma in broader communication settings.

\begin{figure}[t]
    \centering
    \includegraphics[width=0.8\linewidth]{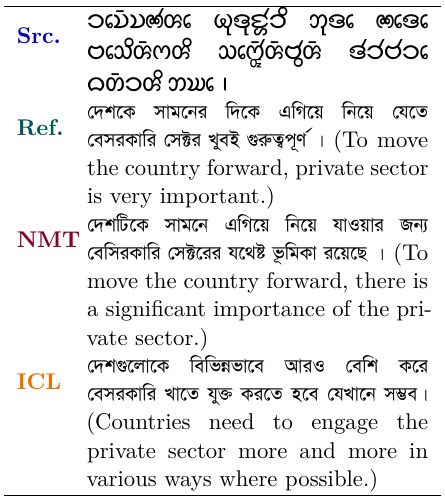}
    \caption{Illustrative Chakma$\rightarrow$Bangla translation example comparing our two best-performing approaches: 
    fine-tuned NMT (BanglaT5) and in-context learning (GPT with random 400 examples). 
Despite similar automatic scores, the outputs differ in lexical choice and interpretation.}
    \label{fig:intro-example}
\end{figure}

Compared to higher-resource Indo-Aryan languages such as Bangla and Hindi, Chakma has received very limited attention in NLP research.
Existing work is largely restricted to character recognition \cite{Podder} and speech language identification \cite{Pratap2023ScalingST}.
In contrast, while state-of-the-art commercial large language models (e.g., GPT and Grok) can recognize Chakma script, they often fail to generate semantically faithful Chakma sentences, limiting their reliability in this low-resource setting (see Section~\ref{sec:qualitative_analysis}).
As a result, foundational text-based capabilities-most notably machine translation (MT), which is critical for cross-lingual communication and access to public resources-remain largely unexplored.

Motivated by this gap, we present the first systematic study of machine translation for Chakma, an extremely low-resource and endangered language.
We investigate how different modeling paradigms, including classical machine translation, neural models, and large language models via in-context learning, behave under severe data scarcity and cross-script conditions.
Our goal is to establish practical baselines for Chakma MT while highlighting challenges that arise in extremely low-resource and non-standardized languages.
Figure~\ref{fig:intro-example} illustrates qualitative differences between our best-performing systems.
A fine-tuned BanglaT5 model produces accurate translations with limited parallel data.
A GPT-based in-context learning approach, using only 400 demonstration examples, generates outputs that remain semantically faithful despite minimal supervision.

\noindent
Our contributions are summarized as follows:

\noindent
(1) \textbf{First Chakma--Bangla MT resources.}
We release the first Chakma--Bangla parallel corpus with 15,021 sentence pairs, 
a large Chakma monolingual corpus with 42,783 sentences, and 
a curated trilingual Chakma--Bangla--English benchmark with 600 evaluation sentences.
These resources provide a foundation for machine translation and downstream NLP research on Chakma.

\noindent
(2) \textbf{Script-bridging transliteration framework.}
We propose a simple, rule-based character-level transliteration system that {leverages Chakma--Bangla script similarity} to provide near one-to-one mappings, preserving semantic content while bridging script differences.
This enables effective cross-script transfer and allows pretrained and large language models to be applied in this extremely low-resource setting.

\noindent
(3) \textbf{Comprehensive benchmarking in an extremely low-resource setting.}
We systematically benchmark statistical and neural MT, fine-tuned pretrained models (e.g., BanglaT5, mT5), and GPT-based in-context learning, establishing strong and robust baselines for Chakma-Bangla translation.

\noindent
(4) \textbf{Analysis of orthographic variability.}
We analyze orthographic inconsistencies in Chakma arising from non-standardized spelling and script usage.
We show that this variability substantially affects automatic evaluation and MT performance, with BLEU underestimating translation quality due to multiple valid spellings.

\begin{table*}[!ht]
\centering
\resizebox{0.9\textwidth}{!}{
\begin{tabular}{llccc}
\toprule
\textbf{Category} & \textbf{Source} & \textbf{Samples} & \textbf{Avg \#Tok} & \textbf{Total} \\
\midrule
\multirow{5}{*}{Parallel}    
& UN-Disabilities (BN-CCP) & 610 & 16.86 & \multirow{5}{*}{15,021} \\
& UN-Child Rights (BN-CCP) & 291 & 37.66 & \\
& Dictionary (word pairs) & 5,473 & 1.14 & \\
& Crowdsourced (BN-EN-CCP) & 3,444 & 4.51 & \\
& Expert translations (BN-EN-CCP) & 5,203 & 3.60 & \\
\midrule
Monolingual & Local Chakma sources (CCP) & 42,783 & 5.81 & 42,783 \\
\midrule
Evaluation & RisingNews Benchmark Extension (BN-EN-CCP) & 600 & 14.8 & 600 \\
\bottomrule
\end{tabular}
}
\caption{Overview of the Chakma--Bangla datasets introduced in this work, including data sources, sample counts, and average sequence length (Avg \#Tok) measured in space-separated Chakma (CCP) tokens; BN and EN denote Bangla and English, respectively.}
\label{tab:dataset}
\end{table*}

\section{Related Works}
\label{sec: related-works}
Machine translation research has historically focused on high-resource language pairs, where large parallel corpora were readily available \citep{six-challenges}.
Early work was dominated by statistical machine translation systems trained on millions of sentence pairs for high-resource language pairs \citep{SMT-base}.
With the advent of neural machine translation, attention-based encoder–decoder models further reinforced this reliance on abundant parallel data \citep{bahdanau}.
As a result, languages with scarce digital resources have received comparatively less attention and continue to face substantial limitations in existing MT systems.

More recent work in low-resource NMT has explored a range of strategies to mitigate data scarcity, including semi-supervised learning with monolingual data \citep{gulcehre}, back-translation \citep{sennrich2016improvingneuralmachinetranslation}, multilingual neural machine translation for cross-lingual transfer \citep{kocmi-bojar-2018-trivial}, and transliteration for closely related languages with different scripts \citep{crossscript_transliteration_mt}, particularly for Asian and Indigenous languages \citep{Riza}.
In extremely low-resource settings, prior work reports very low translation quality overall (often below 10 BLEU) \citep{guzman-etal-2019-flores}, with several MT approaches yielding BLEU scores in the low single digits or around 1–2 under out-of-domain evaluation, highlighting the difficulty of generalization \citep{zhang-etal-2020-chren}.

More recently, large language models have introduced in-context learning (ICL) as an alternative to fine-tuning for low-resource translation \citep{fewshot_llm}. 
While promising, ICL performance is direction-dependent and varies across language pairs, often favoring high-resource target languages \citep{fewshot_llm}.
These limitations motivate us to systematically study the effectiveness of ICL in extremely low-resource and cross-script translation settings.

\section{ChakmaNMT Dataset}
\label{sec:dataset-description}

Table~\ref{tab:dataset} summarizes the parallel, monolingual, and evaluation data collected in this work, with details discussed below.

\subsection{Parallel Corpus}

\paragraph{Parallel Documents}
After extensive searching, we identified two publicly available documents that contain aligned Bangla(BN) and Chakma(CCP) translations: \emph{UN Convention on the Rights of Persons with Disabilities}~\cite{UN} and \emph{UN Convention on the Rights of the Child}~\cite{HumanRights}.  
The Bangla versions of these documents were available only as scanned PDF files containing images of the original printed documents.  
We applied Tesseract OCR\footnote{https://github.com/tesseract-ocr/tesseract} to extract Bangla text from these scans.

Automatic sentence alignment methods such as Hualign~\citep{hualign} were not applicable in our setting due to the lack of a sufficiently rich Chakma lexicon.  
Consequently, all sentence alignments were performed manually.  
This process resulted in 610 and 291 CCP-BN sentence pairs from the two documents, respectively.  
In addition, we incorporated word-level translation pairs from the only available Chakma dictionary, provided to us in JSON format.\footnote{The dictionary data were provided directly by the dictionary’s owner for research use.}
This yielded 5{,}473 additional parallel samples.

\paragraph{Manual Translation by Experts}
To obtain high-quality sentence-level translations, we organized a manual data collection effort involving native Chakma speakers with strong literacy skills.  
We prepared paper-based forms containing 10,000 Bangla sentences randomly sampled from the BN-EN corpus of \citet{hasan}.  
A three-day voluntary translation program was conducted in Dighinala, Khagrachari (Bangladesh).  
Between 7 and 10 proficient Chakma speakers participated each day.  

All collected translations were subsequently reviewed and validated by senior linguistic experts to ensure accuracy and consistency.  
After filtering and quality control, this process produced a final set of 5,203 high-quality CCP-BN-EN parallel sentence triples.  
Further details of this collection process are provided in Appendix~\ref{sec:manual-data-collection}.  

\paragraph{Manual Translation via Crowdsourcing}
To further expand the dataset, we collected additional translations through a crowdsourcing approach involving non-expert Chakma speakers.  
We developed a web-based platform that displayed Bangla sentences and allowed users to submit their Chakma translations.  
The platform link was distributed through social media channels.  

The source sentences primarily consisted of common conversational expressions collected from publicly available resources\footnote{\url{https://www.learnenglishfrombangla.com/2021/07/easily-learn-english-in-bangla-beginner.html}, \url{https://www.omniglot.com/language/phrases/bengali.php}, and \url{https://en.wikibooks.org/wiki/Bengali/Common_phrases}}.  
These resources already include English translations.  
Since most contributors were unfamiliar with the Chakma script, they were instructed to write Chakma using Bangla characters.  
We later converted these submissions into Chakma script using a custom transliteration system (See Section~\ref{subsec:transliteration}) developed for CCP-BN conversion.  
After manual verification and filtering by Chakma language experts, this effort yielded 3,444 additional CCP-BN-EN sentence triples.  

Overall, the parallel data collection process resulted in 15,021 BN-CCP parallel sentence pairs, of which 8,647 include aligned English translations.

\subsection{Monolingual Data}
Figure~\ref{fig:mono-pie-chart} shows the distribution of the collected Chakma monolingual data by content type.
We collected a substantial amount of Chakma monolingual data relative to the available parallel resources.
Due to the scarcity of digitally available Chakma texts,\footnote{Chakma digital texts and scripts are rarely available online, and most existing materials are accessible only through local scholars or printed sources.} we conducted in-person visits to Chakma language scholars to obtain soft copies of Chakma script materials.
These materials primarily consist of poems, articles, short stories, and a small number of national-level textbooks.
In addition, we collected Indian Chakma textbooks and a Chakma folktale mobile application, 
and we reused the Chakma dictionary introduced in the parallel data collection, 
which contains numerous high-quality example sentences accompanying lexical entries and is therefore well suited for monolingual data extraction.

To process these sources, all materials were first transcribed into separate \texttt{.docx} files, which preserved the original Chakma fonts used in the documents.
However, these fonts were encoded in various ASCII-based formats, each with distinct character mappings.
To address this, we developed a conversion program that maps all source fonts to a unified Unicode font, RebangUni,\footnote{\url{https://github.com/Bivuti/RibengUni}}
 the first UTF-8–compliant font for the Chakma language.
This enabled consistent normalization across heterogeneous sources.

Finally, we applied a simple rule-based segmentation procedure, splitting text at sentence boundaries defined by three punctuation markers: ?', !', and `|' (full stop).
After preprocessing and normalization, we obtained a total of 42,783 Chakma monolingual samples.
Tables~\ref{tab:monolingual-details-1} and~\ref{tab:monolingual-details-2} provide detailed descriptions of the monolingual data sources, while the font conversion code is available in our GitHub repository\footnote{\url{https://github.com/Aunabil4602/chakma-nmt-normalizer}} and the list of supported ASCII fonts is provided in Appendix Table~\ref{tab:ascii-font-list}.

\subsection{Evaluation Data}
\label{sec:evaluation-data}
To evaluate our models, we constructed a carefully curated benchmark dataset (see Table~\ref{tab:dataset}).
We randomly selected 500 Bangla--English sentence pairs from the RisingNews Benchmark dataset, which consists of online news articles, introduced by \citet{hasan}.
The RisingNews dataset was preprocessed and filtered following the methodology of \citet{guzman-etal-2019-flores}, making it a high-quality and widely used evaluation resource.
Since the dataset already contains bilingual sentence pairs, translating these sentences into Chakma enables the construction of a trilingual benchmark spanning Bangla, English, and Chakma.
We asked three Chakma language researchers, who had not participated in the parallel data collection, to independently translate these sentences into Chakma.
Each annotator translated 200 sentences, with an overlap of 50 sentences shared across all three annotators.
The shared subset was included to allow analysis of translation variability and orthographic inconsistency across gold references, which we further discuss in Section~\ref{sec:qualitative_analysis}.
In total, this process resulted in 600 evaluation samples, which we refer to as the \textit{RisingNewsChakma} benchmark.
This benchmark is out-of-domain with respect to our training data and is used exclusively for evaluation.

\section{Machine Translation Approaches}
\label{sec:methods}

\paragraph{Machine Translation Task}
This task is formulated as a sequence-to-sequence learning problem at the sentence level.  
Given a source-language sentence, the model generates a target-language sentence token by token.  
The objective is to learn this mapping from extremely limited parallel data.  
We study this setting for machine translation between Chakma and Bangla.

We compare three complementary approaches to establish strong baselines and understand what works best for this language pair.  
Our methods differ primarily in how they leverage prior knowledge and handle the script mismatch between Chakma and Bangla.  
First, we train conventional MT systems from scratch using only our collected parallel data, which are directly limited by data scarcity.  
Second, we fine-tune pretrained sequence-to-sequence models by transferring knowledge from Bangla via script-bridging transliteration and monolingual augmentation.  
Third, we evaluate large language models using few-shot in-context learning, adapting them to Chakma translation without any parameter updates.  

\subsection{From-Scratch MT Baselines}
\label{subsec:from-scratch}

We train both statistical and neural baselines from scratch on our parallel corpus.  
We use phrase-based SMT \citep{SMT-base} as a classical baseline that remains competitive in low-resource scenarios \citep{six-challenges}.  
We also train a GRU-based RNN with attention \citep{bahdanau,luong-etal-2015-effective} as a lightweight neural model.  
Finally, we train a Transformer \citep{attention-is-all-you-need} as a stronger but more data-demanding neural baseline.

\subsection{Script-Bridging Transliteration}
\label{subsec:transliteration}

To enable the use of pretrained sequence-to-sequence and large language models, we develop a rule-based, character-level transliteration system that bridges the Chakma and Bangla scripts in a near one-to-one manner.\footnote{The transliteration code is publicly available on GitHub (\url{https://github.com/Aunabil4602/chakma-nmt-normalizer}).}
The system preserves phonetic and lexical content while mapping Chakma text into the Bangla Unicode range, allowing Chakma data to be directly processed by Bangla-pretrained models.
This transliteration step serves as a core foundation for both fine-tuning pretrained models and in-context learning experiments.

\paragraph{Exploiting a unique relationship between Chakma and Bangla.}
Chakma and Bangla exhibit an unusually high degree of phonetic and orthographic similarity among Indo-Aryan languages, making transliteration a natural and low-effort strategy to bridge script mismatch and enable effective transfer from Bangla-pretrained models.
Beyond script, the languages also share a largely similar subject--object--verb (SOV) word order, which further supports cross-lingual transfer.
At the same time, Chakma has systematic morphosyntactic differences (e.g., placing negation before the verb), which may introduce local reordering effects beyond script-level variation.

The transliteration is largely based on straightforward one-to-one character mappings in both directions, with a small number of deterministic, phonetics-based normalizations only where exact script-level equivalence is unavailable.
The system prioritizes content preservation over strict character reversibility: round-trip transliteration may introduce minor surface-level variation, but does not result in semantic or lexical information loss (Section~\ref{sec:qualitative_analysis}).
In practice, transliteration is used exclusively as preprocessing and postprocessing: if Chakma is the source, input is transliterated into Bangla before translation; if Chakma is the target, Bangla-script model output is transliterated back into Chakma for evaluation.
All pretrained sequence-to-sequence models described in the following subsection are trained and evaluated using this transliterated input--output representation.

We provide full mapping statistics, handling of non-direct characters, and a summary of all non-direct rules (Figure~\ref{fig:missing_mapping}) in Appendix~\ref{app:transliteration_details}.

\subsection{Fine-Tuning Pretrained Sequence-to-Sequence Models}
\label{subsec:fine-tuning}

We fine-tune pretrained text-to-text models on transliterated Chakma--Bangla parallel data.  
We experiment with BanglaT5 \citep{T5}, mT5-small \citep{xue-etal-2021-mt5}, and mBART \citep{mBART}.  
These models allow us to transfer linguistic knowledge from Bangla and multilingual pretraining into the Chakma setting.  

We further evaluate two data-centric extensions to improve robustness under scarcity.  
We apply iterative back-translation (IBT) \citep{iterative-bt} to generate synthetic parallel data from monolingual corpora.  
In IBT, we start with the forward direction CCP$\rightarrow$BN trained on the original parallel data, the backward direction (BN$\rightarrow$CCP) is trained with synthetic data, and in later iterations both directional models are trained with additional synthetic data.
We also evaluate multilingual training (MNMT) following \citet{johnson-etal-2017-googles}.  
For MNMT, we add 10k Bangla--English sentence pairs from \citet{hasan} to the training data to encourage cross-lingual transfer across Bangla, Chakma, and English.

\subsection{Large Language Models via In-Context Learning}
\label{subsec:icl}

We evaluate few-shot in-context learning as an alternative to fine-tuning using state-of-the-art large language models.  
Specifically, we test GPT-4.1, GPT-4.1-mini, and GPT-o4-mini with a fixed prompting template and a limited number of demonstration translation pairs.  
This setting assesses their ability to perform Chakma translation using only in-context examples.
The full prompting template and example format are shown in Figure~\ref{tab:icl_expeirment_prompt}.

\paragraph{Example Selection}
Demonstration examples are selected using two retrieval strategies: uniform random sampling from the parallel data and character-level n-gram similarity with the input sentence.  
We use character-level matching to handle orthographic variation in Chakma, where multiple valid spellings make word-level retrieval unreliable.  
Comparing these strategies allows us to evaluate whether demonstration relevance, beyond the number of examples, improves in-context translation quality.

\section{Experimental Setup}
\label{sec:experimental-setup}

\paragraph{Data Splits and Evaluation}
We split the parallel Chakma--Bangla corpus into training and development sets.\footnote{The dataset is publicly available on Hugging Face at
amlan107/chakma-nmt-complete-dataset.}
The training set contains 12{,}016 sentence pairs and the development set contains 3{,}005 sentence pairs.
We evaluate all models on the RisingNewsChakma benchmark, which is out-of-domain with respect to the training data.
We report BLEU \citep{sacreBleu} and chrF \citep{chrf} as our primary automatic metrics; following common practice, 
BLEU is used for model selection on the development set, and chrF is reported alongside BLEU for all experimental results.

\paragraph{From-Scratch SMT and NMT}
We use the Moses toolkit\footnote{\url{https://www2.statmt.org/moses/}} for phrase-based SMT and PyTorch for neural models.
All neural models are trained on Google Colab using NVIDIA V100/A100 GPUs\footnote{\url{https://colab.research.google.com/}}.
Data preprocessing follows the normalization\footnote{\url{https://github.com/Aunabil4602/chakma-nmt-normalizer}} scheme of \citet{hasan} with minor language-specific adjustments.
We apply SentencePiece \citep{kudo-richardson-2018-sentencepiece} for tokenization across SMT and NMT systems.
Decoding uses beam search with width 5 and a maximum sequence length of 128 tokens.
We train the GRU-based and Transformer models described in Section~\ref{subsec:from-scratch}.

\paragraph{Fine-Tuning and In-Context Learning}
We fine-tune BanglaT5, mT5-small, and mBART for Chakma--Bangla translation.
For iterative back-translation, we use the full Chakma monolingual corpus and 50k Bangla monolingual sentences.

We also evaluate large language models using few-shot in-context learning without parameter updates, including GPT-4.1, GPT-4.1-mini, and GPT-o4-mini.
We use default decoding settings with temperature set to 1 due to budget constraints.
Each prompt includes between 100 and 400 example translation pairs and translates 20 input sentences, selected as a practical compromise between prompt utilization and computational cost.
Demonstration examples are retrieved from the training split of the parallel corpus and selected either randomly or via character-level $n$-gram similarity, with $n=6$ fixed for stability.

We additionally conduct ablation experiments by (i) removing transliteration for fine-tuned and in-context models, and (ii) evaluating a zero-shot in-context learning configuration without demonstrations.

\noindent Additional experimental details necessary for replication, including normalization,
training hyper-parameters, model architectures and initialization,
multilingual data formatting and oversampling,
in-context learning prompt construction, and multiple runs and randomness,
are provided in Appendix~\ref{sec:appendix-experimental-details}.

\begin{table*}[h!]
\centering
\setlength{\tabcolsep}{3pt}
\resizebox{\textwidth}{!}{
\begin{tabular}{l
    r@{ $\pm$ }l r@{ $\pm$ }l r@{ $\pm$ }l r@{ $\pm$ }l
    r@{ $\pm$ }l r@{ $\pm$ }l r@{ $\pm$ }l r@{ $\pm$ }l}
\toprule
\multirow{3}{*}{\textbf{System}} 
 & \multicolumn{8}{c}{\textbf{BN$\rightarrow$CCP}} 
 & \multicolumn{8}{c}{\textbf{CCP$\rightarrow$BN}} \\
\cmidrule(lr){2-9} \cmidrule(lr){10-17}
 & \multicolumn{4}{c}{\textbf{Dev}} & \multicolumn{4}{c}{\textbf{Test}} 
 & \multicolumn{4}{c}{\textbf{Dev}} & \multicolumn{4}{c}{\textbf{Test}} \\
 & \multicolumn{2}{c}{\textbf{BLEU}} & \multicolumn{2}{c}{\textbf{chrF}} & \multicolumn{2}{c}{\textbf{BLEU}} & \multicolumn{2}{c}{\textbf{chrF}}
 & \multicolumn{2}{c}{\textbf{BLEU}} & \multicolumn{2}{c}{\textbf{chrF}} & \multicolumn{2}{c}{\textbf{BLEU}} & \multicolumn{2}{c}{\textbf{chrF}} \\
\midrule
\multicolumn{17}{c}{{From-scratch trained}} \\ \midrule
SMT & 04.6 & $--$ & \textbf{26.3} & $--$ & 00.1 & $--$ & 20.2 & $--$ & 03.2 & $--$ & 24.9 & $--$ & 00.1 & $--$ & \textbf{20.4} & $--$ \\
RNN & \textbf{05.3} & 0.91 & 22.2 & 3.71 & 00.1 & 0.03 & 10.4 & 2.45 & \textbf{08.0} & 2.51 & 20.1 & 4.37 & 00.1 & 0.07 & 08.5 & 2.43 \\
Transformer & 01.6 & 0.03 & 25.8 & 0.21 & \textbf{00.2} & 0.04 & \textbf{23.2} & 0.30 & 03.0 & 0.12 & \textbf{27.2} & 0.28 & \textbf{00.4} & 0.05 & 20.0 & 0.51 \\
\midrule
\multicolumn{17}{c}{{Fine-tuned  with tranliteration enabled}} \\ \midrule
mBART & 03.5 & 0.35 & 20.0 & 1.45 & 00.2 & 0.03 & 11.1 & 1.09 & 10.4 & 6.74 & 22.6 & 9.80 & 00.8 & 0.51 & 12.8 & 3.49 \\
mT5-small & 08.5 & 0.11 & 32.2 & 0.20 & 02.0 & 0.04 & 25.8 & 0.12 & 14.1 & 0.33 & 33.9 & 0.35 & 04.6 & 0.07 & 27.2 & 0.25 \\
\hspace{1em}+IBT-1it & 08.4 & 0.03 & 33.2 & 0.04 & 02.5 & 0.02 & 29.4 & 0.09 & 14.1 & 0.33 & 33.9 & 0.35 & 04.6 & 0.07 & 27.2 & 0.25 \\
\hspace{1em}+IBT-2it & 08.3 & 0.09 & 33.9 & 0.14 & \textbf{02.7} & 0.09 & 30.1 & 0.09 & 15.1 & 0.11 & 37.4 & 0.14 & 07.8 & 0.10 & 37.8 & 0.20 \\
\hspace{1em}+MNMT & 06.1 & 0.14 & 27.5 & 0.35 & 01.5 & 0.05 & 24.1 & 0.73 & 09.2 & 0.28 & 28.1 & 0.30 & 03.0 & 0.10 & 23.5 & 0.34 \\
BanglaT5 & \textbf{10.9} & {0.06} & 36.3 & 0.22 & \textbf{02.7} & {0.05} & \textbf{30.6} & 0.21 & \textbf{24.4} & {0.11} & 44.6 & 0.08 & 11.7 & 0.19 & 37.9 & 0.09 \\
\hspace{1em}+IBT-1it & 10.4 & 0.19 & 36.5 & 0.11 & 02.2 & 0.06 & 27.7 & 0.21 & \textbf{24.4} & {0.11} & 44.6 & 0.08 & 11.7 & 0.19 & 37.9 & 0.09 \\
\hspace{1em}+IBT-2it & 10.5 & 0.18 & \textbf{36.8} & {0.07} & 02.5 & 0.18 & 28.5 & 0.35 & 24.2 & 0.11 & \textbf{46.8} & {0.06} & \textbf{13.9} & 0.20 & \textbf{46.3} & 0.15 \\ 
\hspace{1em}+MNMT & 08.2 & 0.26 & 32.1 & 0.22 & 02.4 & 0.11 & 29.2 & 0.32 & 20.9 & 0.62 & 40.1 & 0.39 & 12.8 & 0.12 & 38.5 & 0.61 \\
\midrule
\multicolumn{17}{c}{{In-context Learning  with tranliteration enabled}} \\ \midrule
GPT-4.1-mini(R) & \multicolumn{2}{c}{-} & \multicolumn{2}{c}{-} & 01.1 & 0.06 & 28.4 & 0.19 & \multicolumn{2}{c}{-} & \multicolumn{2}{c}{-} & 10.9 & 0.19 & 40.0 & 0.58 \\
GPT-4.1(R) & \multicolumn{2}{c}{-} & \multicolumn{2}{c}{-} & \textbf{01.6} & 0.13 & 30.4 & 0.12 & \multicolumn{2}{c}{-} & \multicolumn{2}{c}{-} & 16.5 & 0.12 & 48.2 & 0.35 \\
GPT-o4-mini(R) & \multicolumn{2}{c}{-} & \multicolumn{2}{c}{-} & 01.5 & 0.03 & 29.7 & 0.03 & \multicolumn{2}{c}{-} & \multicolumn{2}{c}{-} & 12.5 & 0.47 & 42.9 & 0.08 \\
GPT-4.1-mini(N) & \multicolumn{2}{c}{-} & \multicolumn{2}{c}{-} & 01.2 & 0.01 & 28.7 & 0.25 & \multicolumn{2}{c}{-} & \multicolumn{2}{c}{-} & 10.5 & 0.54 & 40.4 & 0.47 \\
GPT-4.1(N) & \multicolumn{2}{c}{-} & \multicolumn{2}{c}{-} & 01.5 & 0.10 & \textbf{31.3} & {0.53} & \multicolumn{2}{c}{-} & \multicolumn{2}{c}{-} & \textbf{17.8} & {0.12} & \textbf{49.6} & {0.12} \\
GPT-o4-mini(N) & \multicolumn{2}{c}{-} & \multicolumn{2}{c}{-} & 01.2 & 0.07 & 29.7 & 0.14 & \multicolumn{2}{c}{-} & \multicolumn{2}{c}{-} & 12.2 & 0.27 & 42.7 & 0.48 \\
\bottomrule
\end{tabular}}
\caption{Translation performance on BN$\leftrightarrow$CCP across development and test sets. 
We report mean $\pm$ standard deviation for BLEU and chrF.
Results are shown for from-scratch models, fine-tuned pretrained models (including iterative back-translation(IBT), up to two iterations, and multilingual fine-tunin(MNMT), and GPT-based in-context learning (ICL) with 400 examples using random(R) or n-gram(N) similarity-based sampling.
Overall, GPT-based ICL achieves the strongest performance for CCP$\rightarrow$BN, while BanglaT5 yields the highest BLEU for BN$\rightarrow$CCP.}
\label{table:performance_std}
\end{table*}

\section{Results}
\label{sec:results}

\paragraph{Transliteration is essential for effective modeling}
Transliteration is a prerequisite for leveraging pretrained models and yields substantial gains.
As shown in Table~\ref{table:abl_without_translit}, removing transliteration collapses both BLEU and chrF to near-zero across fine-tuning and in-context learning.
Reintroducing transliteration restores usable scores in both translation directions, showing that script conversion is essential.
The parallel drops in BLEU and chrF indicate a failure at the character level rather than a tokenization issue.

\paragraph{In-context learning is the most effective approach, but directionally asymmetric}
With 400 examples, ICL achieves strong CCP$\rightarrow$BN performance and is competitive with the best fine-tuned systems, where BanglaT5 consistently outperforms mT5 (Table~\ref{table:performance_std} and Table~\ref{table:icl_sampling_swapped_headers}).
In contrast, BN$\rightarrow$CCP performance remains low (around 1--2 BLEU) even with the best prompts, while fine-tuned models continue to perform better.
This reveals a clear directional asymmetry: ICL is substantially more effective when translating into Bangla than into Chakma.
chrF mirrors this pattern, showing much higher character-level overlap for CCP$\rightarrow$BN than for BN$\rightarrow$CCP.

\paragraph{From-scratch models fail to generalize under extreme data scarcity}
From-scratch SMT, RNN, and Transformer models achieve modest dev performance but collapse on the test set, with both BLEU and chrF dropping to near-zero levels (Table~\ref{table:performance_std}).
This degradation holds in both translation directions and indicates a failure to generalize under extreme data scarcity.
The parallel collapse in BLEU and chrF further suggests a strong domain mismatch with the RisingNewsChakma benchmark, where models fail to recover even partial character-level matches.

\paragraph{Back-translation improves performance in most settings}
Back-translation improves performance in most settings, with consistent gains across both BLEU and chrF.
The improvements are strongest for CCP$\rightarrow$BN: for BanglaT5, two iterations increase test BLEU from 11.7 to 13.9 and chrF from 37.9 to 46.3, while mT5-small improves BN$\rightarrow$CCP from 2.0 to 2.7 (Table~\ref{table:performance_std}).
Gains in BN$\rightarrow$CCP are smaller or mixed, although the forward model benefits from additional synthetic data in later IBT steps.
Notably, CCP$\rightarrow$BN scores for BanglaT5 and mT5 remain unchanged after the first IBT iteration, since the backward model is initially trained on the same parallel data.
Overall, chrF closely mirrors BLEU, indicating that back-translation improves character-level fidelity rather than only n-gram overlap.

\paragraph{Bilingual fine-tuning outperforms multilingual training in our setting}
Multilingual training underperforms bilingual fine-tuning across both translation directions.
For both BanglaT5 and mT5-small, adding MNMT reduces test performance on both BLEU and chrF relative to bilingual fine-tuning (Table~\ref{table:performance_std}).
This indicates that the additional English data introduces noise that outweighs any cross-lingual transfer benefits in this low-resource setting.
Table~\ref{table:multilingual_en-ccp} further supports this finding, as EN$\leftrightarrow$CCP results are substantially worse than BN$\leftrightarrow$CCP on both metrics, showing degradation even at the character level.

\paragraph{ICL benefits from relevant demonstrations and careful scaling}
N-gram similarity–based selection consistently matches or slightly outperforms random sampling, particularly for CCP$\rightarrow$BN (Table~\ref{table:icl_sampling_swapped_headers}).
Increasing the number of demonstrations improves performance on both BLEU and chrF up to a point, after which gains plateau and become non-monotonic.
This shows that example relevance matters more than sheer quantity, with chrF mirroring BLEU and indicating improved character-level fidelity rather than just word overlap.

\paragraph{Performance-cost trade-offs between ICL and fine-tuning}
ICL achieves strong CCP$\rightarrow$BN performance on both BLEU and chrF with minimal data, but requires large commercial models and costly prompts.
Fine-tuning is cheaper and more stable, particularly for BN$\rightarrow$CCP where ICL underperforms on both metrics.
As a result, the preferred approach depends on whether one prioritizes peak performance under data scarcity or long-term deployment cost, with ICL’s gains concentrated in CCP$\rightarrow$BN and fine-tuning remaining more reliable for BN$\rightarrow$CCP.

\section{Qualitative Analysis}
\label{sec:qualitative_analysis}

\paragraph{BN$\rightarrow$CCP translation is substantially harder than CCP$\rightarrow$BN}
Across all systems, BN$\rightarrow$CCP performance is substantially lower than CCP$\rightarrow$BN on both BLEU and chrF (Table~\ref{table:performance_std}).
Even the best BN$\rightarrow$CCP results reach only about 2--3 BLEU, while CCP$\rightarrow$BN attains 13--18 depending on the method.
This asymmetry likely arises because pretrained models encode Bangla more effectively, making translation into Bangla easier than into Chakma.
The same gap in chrF confirms that this is a genuine character-level difficulty rather than a BLEU-specific artifact.
Figure~\ref{tab:example-sentences} provides representative BN$\rightarrow$CCP outputs across model families.

\paragraph{BLEU underestimates quality due to lexical and orthographic variation of Chakma Language}
We observe a large divergence between BLEU and chrF that reflects lexical and orthographic variation rather than semantic errors.
BLEU is particularly sensitive to re-transliterated Chakma outputs, where spelling variation introduced by script conversion lowers n-gram overlap without degrading meaning.
This is evident in inter-annotator agreement on 50 shared benchmark sentences (Section~\ref{sec:evaluation-data}), which is only 4.48 BLEU but 38.82 chrF, indicating stable character overlap despite differing word forms.
Figure~\ref{tab:chakma-variation} further illustrates multiple valid spellings for common words, strongly penalizing BLEU.
These patterns, driven by the lack of standardized Chakma orthography, motivate treating chrF as a co-primary metric alongside BLEU.

\paragraph{Our transliteration preserves content but is not character-exact}
Round-trip evaluation shows that transliteration largely preserves the input, although it is not perfectly character-faithful.
After one cycle, scores remain strong (41.55 BLEU / 79.32 chrF for BN$\rightarrow$CCP$\rightarrow$BN and 38.37 BLEU / 79.69 chrF for CCP$\rightarrow$BN$\rightarrow$CCP), indicating only minor character-level drift.
After the second cycle, scores reach near-ceiling levels (97.6--100 BLEU and 99.5--100 chrF; Table~\ref{tab:translit_roundtrip}), reflecting stabilization once non-bijective mappings are resolved.
Overall, residual differences are best explained by a small set of nearest-character (phonetic) substitutions for symbols without exact cross-script counterparts—surface variations that preserve pronunciation and meaning rather than causing substantive information loss.
This interpretation is consistent with downstream MT results (Table~\ref{table:performance_std}), where strong systems maintain relatively high chrF despite lower BLEU, while removing transliteration causes both metrics to collapse (Table~\ref{table:abl_without_translit}).

\paragraph{Zero-shot ICL ablation highlights the need for demonstrations}
As an ablation, we evaluate zero-shot ICL without any demonstrations.
For BN$\rightarrow$CCP, BLEU remains below 1 across models (Table~\ref{table:icl_zero_shot}), indicating that zero-shot ICL is ineffective in this direction.
CCP$\rightarrow$BN performs better even without examples, but still lags behind few-shot ICL.
These results confirm that explicit demonstrations are essential for generating Chakma outputs in this extremely low-resource setting.
chrF is similarly low in zero-shot BN$\rightarrow$CCP, underscoring that models fail to recover even partial character overlaps without examples.

\paragraph{LLM variants show different effectiveness under ICL}
Under identical in-context learning (ICL) setups, GPT-4.1 achieves the strongest overall performance, 
while GPT-o4-mini consistently outperforms GPT-4.1-mini in both BLEU and chrF (Table~\ref{table:icl_sampling_swapped_headers}). 
As these models differ in architecture, capacity, and intended design, we do not attribute the observed differences to any single factor. 
Instead, we report them as an empirical comparison of LLM variants under the same ICL conditions.
The chrF ranking matches BLEU, suggesting that model differences affect both token-level and character-level fidelity.

\section{Conclusion}
This work presents a foundational study on machine translation for Chakma, an extremely low-resource and endangered language. We introduce new datasets and a transliteration-based framework that enables effective use of pretrained models and large language models. Results show that leveraging pretrained models and related high-resource languages substantially outperforms training from scratch, while challenges such as translation asymmetry and orthographic variation remain. Overall, this work establishes strong baselines and a practical foundation for future NLP research on endangered languages.

\clearpage

\section*{Ethics}
This work involves data collection for an endangered and low-resource language with the goal of supporting language preservation and accessibility. 
All human-generated data were collected with informed consent from contributors, who voluntarily participated and expressed support for this research. 
No personally identifiable information was collected, and we do not anticipate any significant risks or harms resulting from this work.

\section*{Limitations}
This work is constrained by the extremely low-resource nature of the Chakma language, which limits the size and diversity of available training data and leads to relatively low automatic evaluation scores, a common challenge in low-resource machine translation. 
Our rule-based transliteration framework enables effective cross-script transfer and preserves phonetic and lexical content, but relies on manually designed mappings and is not strictly character-bijective, resulting in minor surface-level variation for a small number of script-specific distinctions without affecting meaning. 
Automatic metrics such as BLEU may further underestimate translation quality due to orthographic variation and multiple valid spellings in Chakma. 
Ultimately, the primary bottleneck remains data scarcity: future work could benefit substantially from automated web-based data crawling and collection systems to expand Chakma textual resources, as well as from more data-driven transliteration and translation approaches tailored to extremely low-resource and non-standardized languages.

% \section*{Acknowledgments}
% Our work would not have been possible without the help of several individuals and organizations. 
% Particularly, we would like to thank Injeb Chakma, who assisted us with any issues related to the Chakma language. 
% We are thankful to Arjya Mitra Chakma, Shanti Chakma, and The Zabarang Foundation for providing important documents. 
% We would also like to express our gratitude to Bivuti Chakma for helping us with the RibengUni font and additional documents. 
% For translating the benchmark data, we are highly grateful to Sujan Chakma, Shanti Chakma, and Binoy Bikash Talukdar. 
% Moreover, we extend our sincere thanks to all the people who supported our work and voluntarily participated in the translation process both online and offline.

% ``Beyond the Hill" group assisted us in data collection and enabled us to compete in a competition hosted by the ICT Division of Bangladesh, where we won 10,000 BDT. Additionally, Sentien.IO generously granted us \$500 for training the models.

% Bibliography entries for the entire Anthology, followed by custom entries
%\bibliography{anthology,custom}
% Custom bibliography entries only
\bibliography{custom}

\appendix

\section{Appendix}
\label{sec:appendix}

\subsection{Additional Details on Expert-Based Data Collection}
\label{sec:manual-data-collection}

Prior to data collection, we conducted a pre-assessment to evaluate the feasibility of manual translation by volunteer Chakma speakers.  
The assessment revealed that participation and translation quality were highly sensitive to task complexity and time requirements.  
In particular, longer sentences significantly increased cognitive load and annotation time, frequently resulting in incomplete or inaccurate translations.  
These difficulties were exacerbated by the fact that many Bangla or English lexical items are either rarely used in Chakma or lack direct lexical equivalents, often requiring paraphrasing or explanatory reformulation.  
Based on these findings, we constrained source sentences to a length of 2 to 8 words to reduce annotation burden while maintaining sufficient linguistic content.

To mitigate these challenges and reduce annotator fatigue, we deliberately selected short sentences, 
with the probability of sentence selection peaking at 4-5 words and gradually decreasing toward both extremes. 
This distribution reflects an optimal trade-off between linguistic informativeness and annotation feasibility. 
Very short sentences (e.g., one word) were avoided due to limited contextual value, while longer sentences were excluded to minimize translation difficulty and error propagation.

The participants were predominantly young native Chakma speakers with functional bilingual proficiency in Bangla and English but limited formal training in translation. 
Despite their linguistic competence, many participants reported difficulty translating abstract or institutional terms, 
as Chakma remains primarily an oral language with limited standardized written usage. 
Furthermore, expert translators typically provided handwritten translations, 
which were later digitized by trained typists due to limited familiarity with Chakma script typing.

We discuss additional things in Appendix~\ref{sec:chakma-challenges} and \ref{sec:interviews}

\subsection{Origins of Orthographic Variation in Chakma}
\label{sec:chakma-challenges}

Chakma lacks a widely accepted standardized grammar, resulting in substantial variation in spelling and syntactic structure across written sources.
Historically, language experts and local shamans have documented the language using personal conventions without publishing formal grammatical guidelines.
As a result, the same lexical items are often written using multiple valid spellings, leading to pervasive orthographic inconsistency.
Disagreements among scholars have further hindered consensus on standard grammatical rules.
These disagreements are also reflected in differences between Indian and Bangladeshi Chakma scholarly traditions.
Such variability complicates data normalization and automatic evaluation in downstream NLP tasks.

\subsection{Script-Bridging Transliteration: Mapping Coverage and Non-Direct Rules}
\label{app:transliteration_details}

\begin{figure}
    \centering
    \includegraphics[width=0.8\linewidth]{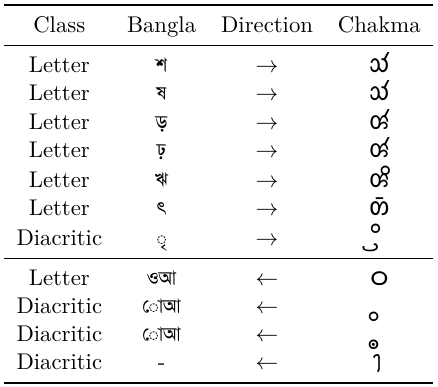}
    \caption{Nearest-character substitutions used in the Chakma–Bangla transliteration system for characters without direct one-to-one mappings. 
These substitutions preserve semantic content and approximate pronunciation, while potentially neutralizing non-contrastive orthographic distinctions. 
The only entry marked with a dash (–) in the Bangla column corresponds to a rare Chakma prosodic lengthening marker that lacks an explicit Bangla graphemic equivalent and is normalized during transliteration. 
All four Chakma characters without direct Bangla counterparts are extremely rare in contemporary usage and have negligible impact on downstream translation quality.}
 \label{fig:missing_mapping}
\end{figure}

This appendix provides the full mapping statistics and the handling rules for characters that do not admit an exact one-to-one correspondence between Chakma (CCP) and Bangla (BN). A summary of all non-direct substitutions in both directions is shown in Figure~\ref{fig:missing_mapping}.

\paragraph{CCP$\rightarrow$BN coverage.}
In the CCP$\rightarrow$BN direction, all 10 Chakma digits have direct mappings.
Among the core Chakma letters (vowels and consonants), 36 out of 37 characters map directly to Bangla equivalents.
For diacritics, 18 out of 21 Chakma diacritical marks have direct mappings.
The remaining Chakma characters correspond to prosodic or orthographic distinctions that do not have explicit representations in Bangla.
Notably, one such character functions as a lengthening/extension marker: it modifies the pronunciation of an adjacent letter rather than introducing independent lexical content.
Since Bangla lacks an equivalent graphemic mechanism for this feature, we normalize it during transliteration by preserving the base character without adding a distinct symbol in the Bangla rendering, thereby maintaining lexical meaning and the underlying pronunciation class.
Moreover, the four Chakma characters without direct Bangla equivalents are extremely rare in contemporary Chakma usage and do not materially affect downstream translation quality.

\paragraph{BN$\rightarrow$CCP coverage.}
In the reverse BN$\rightarrow$CCP direction, all 10 Bangla digits have direct mappings.
Out of 50 Bangla letters, 44 map directly to Chakma characters.
Similarly, 10 out of 11 Bangla diacritics have corresponding Chakma equivalents.
The remaining Bangla characters encode phonetic distinctions that are not contrastive in Chakma orthography and are therefore mapped to the closest Chakma counterparts that best preserve pronunciation and lexical identity.

\paragraph{Deterministic handling of non-direct characters.}
For characters without direct one-to-one correspondences in either direction, we apply deterministic substitutions based on closest phonetic similarity in the target script (Figure~\ref{fig:missing_mapping}).
Consequently, the transliteration system prioritizes content preservation over strict character reversibility: round-trip transliteration may introduce minor surface-level variation, but does not lead to semantic or lexical information loss (Section~\ref{sec:qualitative_analysis}).
Overall, the system remains predominantly a straightforward character mapping scheme, with a small number of rule-based phonetic normalizations applied only when exact script-level equivalence is unavailable.

\subsection{Additional Experimental Details}
\label{sec:appendix-experimental-details}

\paragraph{Multiple Runs and Randomness}
To assess robustness to training stochasticity, we run all experiments three times with different random seeds
(affecting initialization and minibatch order) and report mean and standard deviation.
For in-context learning, we fix the demonstration set and repeat generation three times to quantify decoding variability.
Therefore, reruns under the same configuration are expected to yield scores consistent with the reported mean ± standard deviation.

\paragraph{Normalization details}
On top of the normalization described by \citep{hasan}, we apply a minimal and conservative normalization step uniformly to all text, including training data, model inputs, and output labels, across all experiments.
For Chakma script, we merge a small number of rarely used, phonetically similar vowel variants into a common representation to reduce superficial spelling variation.
This is analogous to collapsing long and short vowel variants in low-resource settings and is intended to simplify orthographic variation while preserving pronunciation and meaning.
We also normalize all bracket symbols to parentheses in order to reduce sparsity in the data.
These normalization steps are applied symmetrically and are not intended to alter semantic content or translation difficulty.

\paragraph{Additional details on metrics}
We report BLEU and chrF scores using \texttt{sacreBLEU} via the Hugging Face \texttt{evaluate} library with default settings: 
BLEU uses a maximum word n-gram order of 4 (BLEU-4), and chrF uses character n-grams of order 6.

\paragraph{SentencePiece Vocabulary Search}
We use SentencePiece \citep{kudo-richardson-2018-sentencepiece} both for (i) vocabulary building and (ii) tokenization for SMT and NMT.
As part of hyper-parameter optimization, we evaluate vocabulary sizes of 1{,}000, 2{,}000, 5{,}000, 10{,}000, and 20{,}000.

\paragraph{Training Hyper-Parameters and Optimization Settings}
We apply gradient clipping with max norm 1.0.
We tune learning rates in \{0.001, 0.005, 0.0001, 0.0005\}, batch sizes in \{8, 16, 32\}, and training steps in \{10{,}000, 15{,}000, 20{,}000\}.
Warmup steps are varied in \{0, 2000, 4000\}.
We also tune label smoothing over \{0.1, 0.2, 0.3, 0.4, 0.5\}.
The final tuned hyper-parameter configurations are reported in Table~\ref{tab:hp-ft}.

\paragraph{RNN with Attention: Architecture and Initialization}
For the RNN baseline, we use a public PyTorch seq2seq implementation.\footnote{\url{https://github.com/bentrevett/pytorch-seq2seq/tree/main}}
The model incorporates Luong-style attention \citep{luong-etal-2015-effective}.
We experiment with 1, 2, and 4 stacked recurrent layers, and consider hidden size and embedding size in \{512, 1024\}.
Dropout is tuned in \{0.1, 0.2, 0.3\}.
All RNN parameters are initialized from a normal distribution with mean 0 and standard deviation 0.1.

\paragraph{Transformer: Model Variants and Initialization}
For Transformer training, we follow the standard Transformer formulation \citep{attention-is-all-you-need}.
We explore MarianNMT-style Transformer\footnote{https://huggingface.co/docs/transformers/model\_doc/marian} configurations available through HuggingFace implementations, and initialize weights with Glorot initialization \citep{glorot}.
We vary the number of layers in \{1, 2, 6\}, attention heads in \{1, 2, 6\}, dropout in \{0.1, 0.2, 0.3\}, and feed-forward hidden dimensions in \{512, 1024\}.

\paragraph{Multilingual Formatting, and Oversampling}
In a multilingual training(MNMT), we prepend a target-language prefix tag to each input sentence: <BN> for Bangla, <EN> for English, and <CCP> for Chakma.
To mitigate imbalance, we oversample Chakma-involving parallel pairs to better balance all translation directions, following practices shown to improve multilingual performance \citep{johnson-etal-2017-googles}.

\paragraph{In-Context Learning Prompt Construction}
For in-context learning (ICL), demonstration examples are selected either uniformly at random or using character-level $n$-gram overlap to account for orthographic variation in Chakma.
Character-level matching is used instead of word-level matching due to the absence of standardized spelling.
The $n$-gram size was selected through limited manual experimentation using a small subset of the development data,
due to computational budget constraints.
This subset was used only for preliminary testing of retrieval behavior, and not for model selection or final evaluation.
We evaluated a narrow range of values and fixed $n=6$, which provided the most stable retrieval behavior in these tests.

\paragraph{mBART: Additional Details}
We do not apply iterative back-translation (IBT) or multilingual fine-tuning (MNMT) to mBART, as its plain fine-tuning performance is substantially lower than other models (Table~\ref{table:performance_std}, Section~\ref{sec:results}), making these extensions unlikely to provide meaningful improvements.

\subsection{Interviews with Chakma Language Experts}
\label{sec:interviews}
We interviewed several scholars in Bangladesh to discuss the variants, for example, the number of characters, diacritics, rules, spelling patterns, etc. The scholars include Arjya Mitra, Injeb Chakma, Ananda Mohon Chakma, and Sugata Chakma. However, almost all of them suggested following the rules maintained by the members of the National Curriculum and Textbook Board of Bangladesh involved in writing the Chakma books for the pre-primary levels because their rules will be followed eventually. The most important rule from them that we followed in our transliteration codes from Bangla to Chakma, is that the core grapheme cannot have more than one diacritic attached to a consonant or a vowel. However, in India, this restriction is not maintained, rather more than one diacritic is seen frequently in their documents. 

\begin{figure}[th]
    \centering
    \begin{tikzpicture}
        \pie[
            radius=1.2,
            explode=0.1,
            text=legend,
            hide number,
            color={
                blue!60,
                teal!60,
                orange!70,
                purple!55,
                green!60,
                red!60,
                gray!60
            }
        ]
        {34.1/Dictionary 34.1\%, 
         22.2/Story 22.2\%, 
         20.5/Textbook 20.5\%, 
         10.8/Poem 10.8\%, 
         10.5/Novel 10.5\%,  
         1/Article 1\%, 
         0.9/Other 0.9\%}
    \end{tikzpicture}
    \caption{Distribution of Chakma monolingual data by content type. 
    The corpus contains 42,783 monolingual samples collected from diverse sources, including dictionaries, stories, textbooks, poems, novels, and articles.}
    \label{fig:mono-pie-chart}
\end{figure}

\begin{table}
\centering
\resizebox{\columnwidth}{!}{
\begin{tabular}{lcc}
\toprule
\textbf{Title} & \textbf{Content} & \textbf{Samples} \\
\midrule
Ajanir dajan firana.docx & Story & 206 \\
Amader-Bari-2.pdf & Story & 12 \\
Amader-Bari-3.pdf & Story & 23 \\
Amader-gaye-dewar-pinon.pdf & Story & 10 \\
Amar-Charar-Boi.pdf & Poem & 123 \\
Amlokir-Gach.pdf & Story & 27 \\
Article 3rd Jamachug.docx & Story & 194 \\
Article 4th Furamon.docx & Story & 194 \\
Article 5th Pawr Murah.docx & Story & 191 \\
Bang-O-Puti-mach.pdf & Story & 11 \\
Banor-Berate-Eseche.pdf & Story & 35 \\
Banorer-Marfa-khaowa.pdf & Story & 10 \\
Bashir-soor.pdf & Story & 9 \\
Bie-Bari.pdf & Story & 28 \\
Bijhu.pdf & Story & 28 \\
Binoy Bikash Talukder20.docx & Poem & 647 \\
Binoy Dewan.docx & Poem & 2004 \\
Bizute-Berano.pdf & Story & 12 \\
Bone-Gie-Gach-Kata.pdf & Story & 30 \\
Boner-Mama.pdf & Story & 11 \\
Chader-Buri.pdf & Story & 28 \\
Chakma Dictionary app & Other & 14928 \\
Chakma Folktales app & Story & 3765 \\
Chakma Love song Uvagit.docx & Story & 13 \\
Chakma Text Book For Class-IV 2010 (IN Govt).docx & Textbook & 1088 \\
Chakma Text Book for Class-II  2010 (IN Govt).docx & Textbook & 490 \\
Chakma Text Book for Class-III  2010 (IN Govt).docx & Textbook & 561 \\
Chakma Text Book for Class-V  2010 (IN Govt).docx & Textbook & 940 \\
Chakma Text Book for Class-VI 2010 (IN Govt).docx & Textbook & 1543 \\
Chakma Text Book for Class-VII 2010 (IN Govt).docx & Textbook & 1858 \\
Chakma.docx & Article & 136 \\
Charar Boi-Chakma-Pages.pdf & Poem & 31 \\
Cijir Orago Boi-Chakma-Pages.pdf & Other & 71 \\
Cijir Talmiloni Kodatara-Chakma-Pages.pdf & Other & 45 \\
Cycle-e-Bazare-Jawa.pdf & Story & 33 \\
Dhanpudi.doc & Story & 1278 \\
Dudur-Kanna.pdf & Story & 40 \\
Dui-Bandhobir-Kotha.pdf & Story & 16 \\
Ghara Poja pire-Chakma-Pages.pdf & Other & 4 \\
H.F.Miller's Rangakura.docx & Story & 90 \\
Hotat-Agun.pdf & Story & 12 \\
Iskulo Akto-Chakma-Pages.pdf & Other & 5 \\
Jhimit-Ekhon-Bhalo.pdf & Story & 42 \\
Jhogra-Kora-Valo-Noi.pdf & Story & 42 \\
Kalo-and-Forshar-Kotha-1.pdf & Story & 22 \\
Kanamachi-Khela.pdf & Story & 13 \\
Karo-bipode-hasa-thik-na.pdf & Story & 15 \\
Kolar-Kotha-1.pdf & Story & 11 \\
Korgosher-sobji-bagan.pdf & Story & 12 \\
Lairang-er-nodi-par-howa.pdf & Story & 13 \\
Lao-er-Desh-Vromon.pdf & Story & 44 \\
Laz-kata-Banor.pdf & Story & 12 \\
Lobh-kora-valo-na.pdf & Story & 16 \\
\bottomrule
\end{tabular}}
\caption{Names of the sources of our Chakma monolingual data with details (Part 1).}
\label{tab:monolingual-details-1}
\end{table}

\begin{table}
\centering
\resizebox{\columnwidth}{!}{
\begin{tabular}{lcc}
\toprule
\textbf{Title} & \textbf{Content} & \textbf{Samples} \\
\midrule
Mamar-Bari.pdf & Story & 19 \\
Mayer-Upadesh-1.pdf & Story & 19 \\
Meghla-Akash.pdf & Story & 22 \\
Mitar-Fuler-Bagan-1.pdf & Story & 10 \\
Moina-Pakhi-1.pdf & Story & 16 \\
Monar Sabon-Chakma-Pages.pdf & Story & 36 \\
Moni-Malar-Kotha-.pdf & Story & 22 \\
Monir-shopno-dekha.pdf & Story & 14 \\
Morog-Jhuti-Fool.pdf & Story & 25 \\
My Legha by Injeb Chakma.doc & Story & 727 \\
Nada-bhet-math for class I (IN Govt Tripura).docx & Textbook & 878 \\
Nanarakam-ghor.pdf & Story & 14 \\
Nirapod-pani-pan-korbo.pdf & Story & 13 \\
Ojhapador Chora-Chakma-Pages.pdf & Poem & 30 \\
Paka-Lichu.pdf & Story & 19 \\
Porichoy.pdf & Story & 16 \\
Projapoti-Ronger-Kotha.pdf & Story & 12 \\
Puti-Macher-Fal.pdf & Story & 13 \\
Rangdhanu.pdf & Story & 20 \\
Ranjuni for Class I (IN Govt) Tripura.docx & Textbook & 1459 \\
SRM 1st P. Bargang.docx & Poem & 156 \\
SRM 1st R. Krisnachura.docx & Poem & 149 \\
SRM 2nd P. Belwa Pawr.docx & Poem & 259 \\
SRM 2nd R. Chadarok.docx & Poem & 76 \\
Sanye-Pidhe-.pdf & Story & 6 \\
Shikkha Boi2017.docx & Poem & 722 \\
Shing-Macher-Kata.pdf & Story & 36 \\
Shiyal-er-Khang-Garang-Bazano.pdf & Story & 19 \\
Shrout.pdf & Story & 8 \\
Sial-mamar-school.pdf & Story & 14 \\
Sukorer-pat-batha-1.pdf & Story & 12 \\
Surjyer-Manush.pdf & Story & 21 \\
Tanybi.doc & Story & 79 \\
Tarum A Ranjuni-Chakma-Pages.pdf & Other & 16 \\
Teen-bondhur-golpo.pdf & Story & 13 \\
Text-Book-Chakma-pdf.pdf & Story & 1405 \\
Thurong-Barite-Raja.pdf & Story & 43 \\
Tin-bondhur-gacher-kotha.pdf & Story & 15 \\
Tiya-Pakhi-1.pdf & Story & 23 \\
chakma novel  hlachinu.docx & Novel & 1571 \\
chedon akkan(10).pdf & Article & 103 \\
diarrhea-hole-ki-Korbo.pdf & Article & 18 \\
ghila khara  class 3 p. 62.docx & Story & 133 \\
kajer-Kotha.pdf & Story & 11 \\
kochpanar rubo rega.docx & Story & 151 \\
mle- 2 ananda babu.docx & Poem & 174 \\
tin fagala-1.docx & Novel & 1765 \\
Changma Ekbacchya Kodha2.doc & Other & 170 \\ %চাঙমা একবাচ্যা কধা২.docx
Chadi 2 Pojhodhe.docx & Novel & 1209 \\ %চাদি ২ পয়ধে.docx
\bottomrule
\end{tabular}}
\caption{Names of the sources of our Chakma monolingual data with details (Part 2).}
\label{tab:monolingual-details-2}
\end{table}

\begin{table}
\centering
\begin{tabular}{l}
    \toprule
    ASCII Font list of Chakma\\ \midrule
    BivunabaKhamaC \\ 
    BijoygiriDPC \\ 
    Udoy Giri \\ 
    Alaam \\ 
    Arjyaban \\ 
    Chakma(SuJoyan) \\ 
    Punong Jun \\ \bottomrule
\end{tabular}
\caption{Chakma ASCII fonts identified in our data sources and subsequently converted to the RibengUni (UTF-8) font as part of corpus normalization.}
\label{tab:ascii-font-list}
\end{table}

\begin{figure*}
    \centering
    \includegraphics[width=\linewidth]{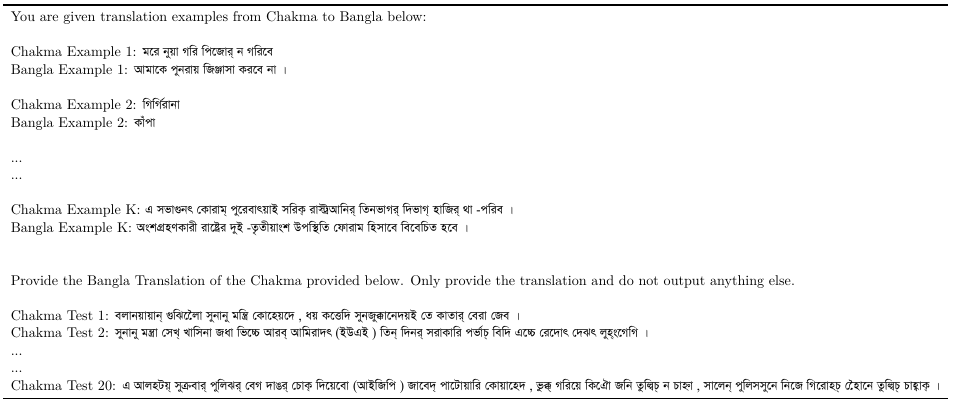}
    % \todo{Add that don't use internet}
    \caption{Prompt format used for our few-shot in-context learning (ICL) experiments, illustrating the structure of source-target examples, task instructions, and test-time input.}
    \label{tab:icl_expeirment_prompt} 
\end{figure*}

\begin{table*}[h!]
\centering
\setlength{\tabcolsep}{3pt}
\resizebox{\textwidth}{!}{
\begin{tabular}{l
    r@{ $\pm$ }l r@{ $\pm$ }l r@{ $\pm$ }l r@{ $\pm$ }l
    r@{ $\pm$ }l r@{ $\pm$ }l r@{ $\pm$ }l r@{ $\pm$ }l}
\toprule
\multirow{3}{*}{\textbf{System}} 
 & \multicolumn{8}{c}{\textbf{BN$\rightarrow$CCP}} 
 & \multicolumn{8}{c}{\textbf{CCP$\rightarrow$BN}} \\
\cmidrule(lr){2-9} \cmidrule(lr){10-17}
 & \multicolumn{4}{c}{\textbf{Dev}} & \multicolumn{4}{c}{\textbf{Test}} 
 & \multicolumn{4}{c}{\textbf{Dev}} & \multicolumn{4}{c}{\textbf{Test}} \\
 & \multicolumn{2}{c}{\textbf{BLEU}} & \multicolumn{2}{c}{\textbf{chrF}} & \multicolumn{2}{c}{\textbf{BLEU}} & \multicolumn{2}{c}{\textbf{chrF}}
 & \multicolumn{2}{c}{\textbf{BLEU}} & \multicolumn{2}{c}{\textbf{chrF}} & \multicolumn{2}{c}{\textbf{BLEU}} & \multicolumn{2}{c}{\textbf{chrF}} \\ \midrule
BanglaT5 & \textbf{10.9} & {0.06} & \textbf{36.3} & 0.22 & \textbf{02.7} & {0.05} & \textbf{30.6} & 0.21 & \textbf{24.4} & {0.11} & \textbf{44.6} & 0.08 & \textbf{11.7} & 0.19 & \textbf{37.9} & 0.09 \\
BanglaT5$^{\mathbf{\dagger}}$ & 00.0 & 0.02 & 00.5 & 0.04 & 00.0 & 0.00 & 00.5 & 0.18 & 02.3 & 0.04 & 14.5 & 0.09 & 00.1 & 0.02 & 14.3 & 0.37 \\ \midrule
mT5-small & \textbf{08.5} & 0.11 & \textbf{32.2} & 0.20 & \textbf{02.0} & 0.04 & \textbf{25.8} & 0.12 & \textbf{14.1} & 0.33 & \textbf{33.9} & 0.35 & \textbf{04.6} & 0.07 & \textbf{27.2} & 0.25 \\
mT5-small$^{\mathbf{\dagger}}$ & 00.2 & 0.17 & 09.1 & 0.91 & 00.0 & 0.00 & 07.7 & 0.23 & 00.8 & 0.04 & 11.9 & 0.99 & 00.1 & 0.00 & 11.7 & 0.37 \\ \midrule
GPT-4.1(R) & \multicolumn{2}{c}{-} & \multicolumn{2}{c}{-} & \textbf{01.6} & 0.13 & \textbf{30.4} & 0.12 & \multicolumn{2}{c}{-} & \multicolumn{2}{c}{-} & \textbf{16.5} & 0.12 & \textbf{48.2} & 0.35 \\
GPT-4.1(R)$^{\mathbf{\dagger}}$ & \multicolumn{2}{c}{-} & \multicolumn{2}{c}{-} & 00.4 & 0.12 & 18.3 & 0.12 & \multicolumn{2}{c}{-} & \multicolumn{2}{c}{-} & 00.3 & 0.10 & 18.2 & 0.21 \\
\bottomrule
\end{tabular}}
\caption{Ablation study comparing models evaluated \emph{with} and \emph{without} transliteration on BN$\rightarrow$CCP and CCP$\rightarrow$BN translation. 
Rows marked with $\dagger$ indicate evaluation \textbf{without transliteration} (native Chakma script as input and output). 
Comparison includes BanglaT5 and mT5-small fine-tuned models, as well as GPT-based in-context learning (ICL) models with random sampling of 400 examples. 
Metrics report mean $\pm$ std for BLEU and chrF on the development and test sets.
Removing transliteration results in near-zero performance across both fine-tuning and ICL, highlighting its necessity in this setting.}
\label{table:abl_without_translit}
\end{table*}

\begin{table}
    \centering
    \resizebox{\columnwidth}{!}{
    \begin{tabular}{l c c c}
        \toprule
            Parameter & RNN & Trans. & T5\\
        \midrule
            Max Epochs & - & - & 5 \\
            Max Train Steps & 20000 & 20000 & - \\
            Warmup Steps/Ratio & 4000 & 4000 & 0.1 \\
            Learning Rate & 0.0005 & 0.0001 & 0.0005 \\
            Batch Size & 16 & 32 & 16\\
            Max Length & 128 & 128 & 128 \\
            Optimizer & adam & adam & adam \\
            Vocab size & 2000 & 10000 & - \\
            Beam width & 5 & 5 & 5 \\ 
            Clip gradient & 1.0 & 1.0 &  - \\
            Label Smoothing & 0.2 & 0.5 & 0.3 \\
            d\_model & - & 512 & - \\
            dropout & - & 0.2 & - \\
            layer\_dropout & - & 0.1 & - \\
            att\_heads & - & 1 & - \\
            ffn\_dim & - & 512 & - \\
            blocks & - & 6 & - \\
            rnn\_dropout & 0.3 & - & - \\
            layer\_normalization & True & - & - \\
            layers & 1 & 6 & - \\
            word\_embedding & 512 & - & - \\
            hidden\_embedding & 1024 & - & - \\
            weight\_decay & - & - & 0.01\\
        \bottomrule
    \end{tabular}}
    \caption{Final training hyperparameters selected based on validation performance for from-scratch RNN and Transformer models, and for fine-tuning pretrained T5-based models (BanglaT5 and mT5-small). mBART was fine-tuned using the same hyperparameter settings as T5.}
    \label{tab:hp-ft}
\end{table}

\begin{table*}
\centering
\setlength{\tabcolsep}{0.05in}
\resizebox{\textwidth}{!}{
\begin{tabular}{l c
    r@{ $\pm$ }l r@{ $\pm$ }l r@{ $\pm$ }l r@{ $\pm$ }l
    r@{ $\pm$ }l r@{ $\pm$ }l r@{ $\pm$ }l r@{ $\pm$ }l}
\toprule
\multirow{3}{*}{\textbf{System}} & \multirow{3}{*}{\textbf{\#Ex.}} 
& \multicolumn{8}{c}{\textbf{Random Sampling}} 
& \multicolumn{8}{c}{\textbf{N-gram Similarity Sampling}} \\
\cmidrule(lr){3-10} \cmidrule(lr){11-18}
&  & \multicolumn{4}{c}{\textbf{BN$\rightarrow$CCP}} & \multicolumn{4}{c}{\textbf{CCP$\rightarrow$BN}}
& \multicolumn{4}{c}{\textbf{BN$\rightarrow$CCP}} & \multicolumn{4}{c}{\textbf{CCP$\rightarrow$BN}} \\
&  & \multicolumn{2}{c}{BLEU} & \multicolumn{2}{c}{chrF}
   & \multicolumn{2}{c}{BLEU} & \multicolumn{2}{c}{chrF}
   & \multicolumn{2}{c}{BLEU} & \multicolumn{2}{c}{chrF}
   & \multicolumn{2}{c}{BLEU} & \multicolumn{2}{c}{chrF} \\
\midrule

      GPT-4.1        & 100 & 01.3 & 0.13	 & 28.7 & 0.40 & 16.2 & 0.23	 & 47.8 & 0.12 & 01.4 & 0.17 & 	30.4 & 0.30 & 16.9 & 0.31	 & 48.7 & 0.28 \\
             & 200 & 01.3 & 0.07	 & 29.5 & 0.20 & \textbf{16.8} & {0.53} & 	47.8 & 0.14 & 01.4 & 0.15	 & 30.5 & 0.35 & 17.5 & 0.28	 & 49.1 & 0.34 \\
             & 400 & \textbf{01.6} & {0.13}	 & \textbf{30.4} & {0.12} & 16.5 & 0.12	 & \textbf{48.2} & {0.35} & \textbf{01.5} & {0.10}	 & \textbf{31.3} & {0.53} & \textbf{17.8} & {0.12}	 & \textbf{49.6} & {0.12} \\
\midrule
       GPT-4.1-mini   & 100    & 00.9 & 0.05 & 27.6 & 0.26 & 10.3 & 0.37 & 40.2 & 0.27 & \textbf{01.2} & 0.10 & 28.6 & 0.35 & \textbf{11.1} & 0.50 & \textbf{40.7} & 0.30 \\
             & 200  & \textbf{01.1} & 0.14 & 28.1 & 0.31 & \textbf{10.9} & 0.20 & \textbf{40.6} & 0.12 & 01.1 & 0.02 & 28.5 & 0.24 & \textbf{11.1} & 0.03 & \textbf{40.7} & 0.10 \\
             & 400 & \textbf{01.1} & 0.06 & \textbf{28.4} & 0.19 & \textbf{10.9} & 0.19 & 40.0 & 0.58 & \textbf{01.2} & 0.01 & \textbf{28.7} & 0.25 & 10.5 & 0.54 & 40.4 & 0.47 \\
\midrule
   GPT-o4-mini          & 100 & 01.3 & 0.04	 & 28.6 & 0.29 & \textbf{12.8} & 0.06	 & 42.7 & 0.03 & 01.1 & 0.05  & 29.2 & 0.14 & \textbf{12.6} & 0.42 & 	\textbf{42.8} & 0.29 \\
             & 200 & 01.5 & 0.02& 	28.3 & 0.51 & \textbf{12.8} & 0.21	 & \textbf{42.9} & 0.16 & 01.4 & 0.04	 & \textbf{29.8} & 0.18 & 12.4 & 0.43 & 	42.2 & 0.60 \\
             & 400 & 01.5 & 0.03 &	\textbf{29.7} & 0.03 & 12.5 & 0.47 & 	\textbf{42.9} & 0.08 & 01.2 & 0.07	 & 29.7 & 0.14 & 12.2 & 0.27 & 	42.7 & 0.48 \\
\bottomrule
\end{tabular}}
\caption{In-context learning (ICL) performance of different GPT variants under identical experimental settings.
Results compare random and n-gram similarity-based sampling of in-context examples across varying numbers of demonstrations (\#Ex.).
BLEU and chrF are reported as mean $\pm$ standard deviation for BN$\rightarrow$CCP and CCP$\rightarrow$BN translation.
The table enables a controlled comparison of LLM variants under the same ICL conditions.
Overall, ICL performance improves as the number of in-context examples increases, with gains becoming more consistent at 200-400 demonstrations across models, 
and n-gram similarity-based sampling generally yielding stronger results than random selection.
}
\label{table:icl_sampling_swapped_headers}
\end{table*}

\begin{table*}
\centering
\setlength{\tabcolsep}{0.07in}
\resizebox{\textwidth}{!}{
\begin{tabular}{l
    r@{ $\pm$ }l r@{ $\pm$ }l
    r@{ $\pm$ }l r@{ $\pm$ }l
    r@{ $\pm$ }l r@{ $\pm$ }l
    r@{ $\pm$ }l r@{ $\pm$ }l}
\toprule
\multirow{2}{*}{\textbf{System}} 
 & \multicolumn{4}{c}{\textbf{EN$\rightarrow$CCP}} 
 & \multicolumn{4}{c}{\textbf{CCP$\rightarrow$EN}}
 & \multicolumn{4}{c}{\textbf{BN$\rightarrow$CCP}}
 & \multicolumn{4}{c}{\textbf{CCP$\rightarrow$BN}} \\
\cmidrule(lr){2-5} \cmidrule(lr){6-9} \cmidrule(lr){10-13} \cmidrule(lr){14-17}
 & \multicolumn{2}{c}{\textbf{BLEU}} & \multicolumn{2}{c}{\textbf{chrF}}
 & \multicolumn{2}{c}{\textbf{BLEU}} & \multicolumn{2}{c}{\textbf{chrF}}
 & \multicolumn{2}{c}{\textbf{BLEU}} & \multicolumn{2}{c}{\textbf{chrF}}
 & \multicolumn{2}{c}{\textbf{BLEU}} & \multicolumn{2}{c}{\textbf{chrF}} \\
\midrule
BanglaT5 & {01.2} & {0.06} & {23.0} & {0.28} & {06.5} & {0.36} & {28.5} & {0.32} & {02.4} & {0.11} & {29.2} & {0.32} & {12.8} & {0.12} & {38.5} & {0.61} \\
mT5-small & 00.2 & 0.01 & 11.1 & 0.88 & 01.0 & 0.18 & 15.6 & 0.34 & 01.5 & 0.05 & 24.1 & 0.73 & 03.0 & 0.10 & 23.5 & 0.34 \\
\bottomrule
\end{tabular}}
\caption{Test-set performance of multilingual fine-tuned models on EN$\leftrightarrow$CCP translation using BanglaT5 and mT5-small. 
BLEU and chrF are reported as mean $\pm$ standard deviation. 
Results for BN$\leftrightarrow$CCP are shown for reference to contrast multilingual performance with the bilingual setting.}
\label{table:multilingual_en-ccp}
\end{table*}

\begin{figure*}
    \centering
    \includegraphics[width=\linewidth]{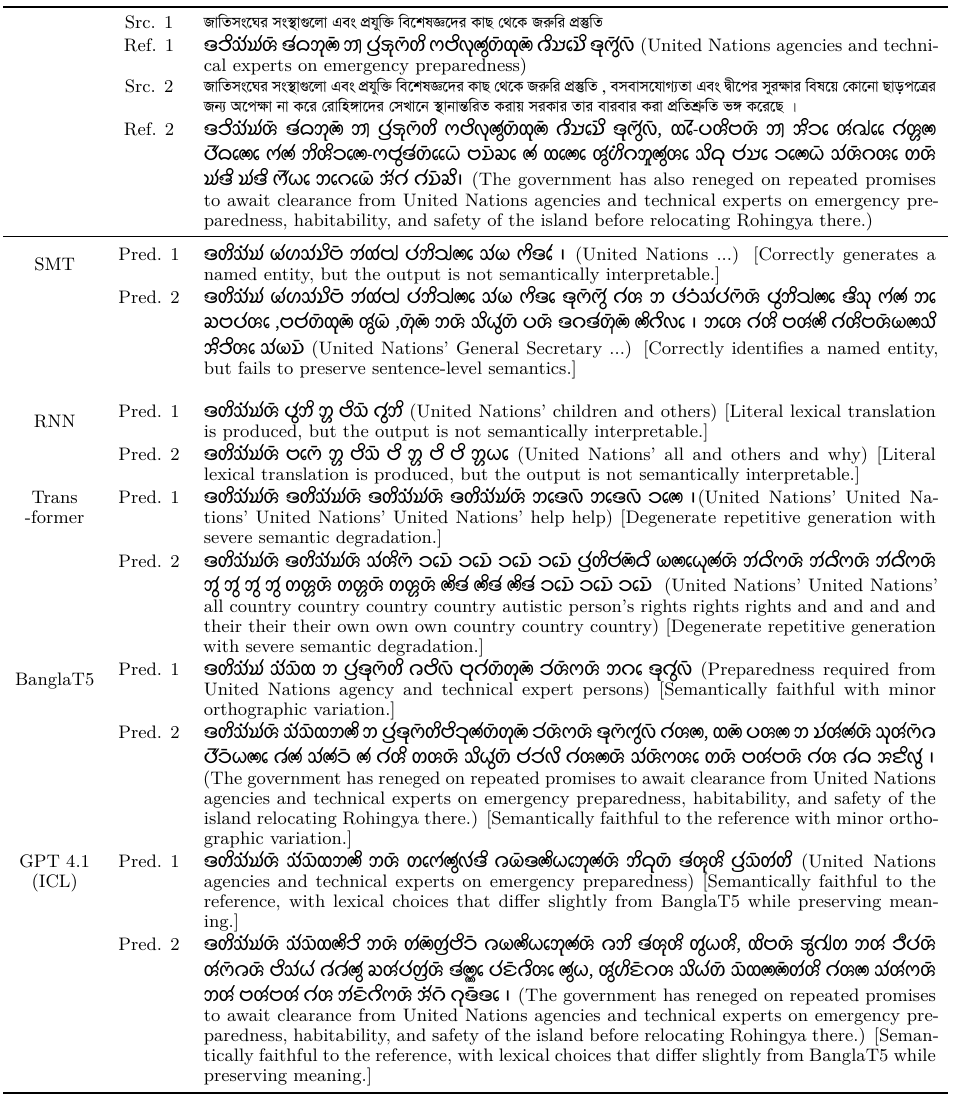}
\caption{Qualitative examples of BN$\rightarrow$CCP translations generated by different models on short and long sentences, illustrating differences in semantic adequacy and robustness across approaches. Parentheses () provide a literal English gloss of each model output for readability, while square brackets [] give a brief qualitative analysis comparing translation quality. 
Overall, BanglaT5 and GPT models using in-context learning (ICL) produce more semantically faithful translations than earlier SMT, RNN, and baseline Transformer approaches, particularly for longer sentences.
}
\label{tab:example-sentences}
\end{figure*}

\begin{table*}
\centering
\setlength{\tabcolsep}{0.08in}
\begin{tabular}{c c c c c}
\toprule
\multirow{2}{*}{\textbf{Round}} 
 & \multicolumn{2}{c}{\textbf{BN$\rightarrow$CCP$\rightarrow$BN}}
 & \multicolumn{2}{c}{\textbf{CCP$\rightarrow$BN$\rightarrow$CCP}} \\
\cmidrule(lr){2-3} \cmidrule(lr){4-5}
 & \textbf{BLEU} & \textbf{chrF} & \textbf{BLEU} & \textbf{chrF} \\
\midrule
1 & 41.55 & 79.32 & 38.37 & 79.69 \\
2 & 99.97 & 99.99 & 97.61 & 99.46 \\
3 & 100.00 & 100.00 & 100.00 & 100.00 \\
\bottomrule
\end{tabular}
\caption{Round-trip transliteration quality up to the third iteration on the Benchmark set. Scores are reported as BLEU and chrF on benchmark sentences and show convergence after two rounds.}
\label{tab:translit_roundtrip}
\end{table*}

\begin{figure*}
    \centering
    \includegraphics[width=0.9\linewidth]{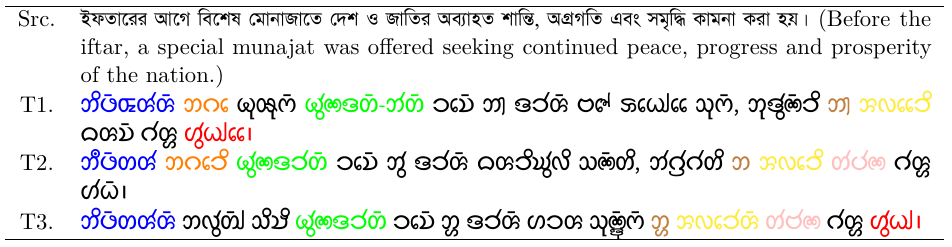}
    \caption{Illustration of orthographic variation in Chakma, where multiple valid spellings of the same word appear across translations produced independently by three language experts from the same Bangla source. Identical colors highlight variant spellings of the same lexical item.}
    \label{tab:chakma-variation}
\end{figure*}

\begin{table*}
\centering
\setlength{\tabcolsep}{0.06in}
\resizebox{0.6\textwidth}{!}{
\begin{tabular}{l
    r@{ $\pm$ }l r@{ $\pm$ }l
    r@{ $\pm$ }l r@{ $\pm$ }l}
\toprule
\multirow{2}{*}{\textbf{System}} 
 & \multicolumn{4}{c}{\textbf{BN$\rightarrow$CCP}} 
 & \multicolumn{4}{c}{\textbf{CCP$\rightarrow$BN}} \\
\cmidrule(lr){2-5} \cmidrule(lr){6-9}
 & \multicolumn{2}{c}{BLEU} & \multicolumn{2}{c}{chrF} 
 & \multicolumn{2}{c}{BLEU} & \multicolumn{2}{c}{chrF} \\
\midrule
{\small GPT-4.1 }     & 00.5 & 0.09	 & 21.1 & 1.44 & {15.5} & {0.49}	 & {45.9} & {0.45} \\
{\small GPT-4.1-mini } & {00.6} & {0.06} & 	{24.2} & {0.71} & 09.2 & 0.13 & 	38.9 & 0.02 \\
{\small GPT-o4-mini }  & {00.6} & {0.03}	 & 22.6 & 1.41 & 12.2 & 0.26 & 	42.2 & 0.23 \\
\bottomrule
\end{tabular}}
\caption{Zero-shot in-context learning ablation showing translation performance with no in-context examples for GPT-4.1, GPT-4.1-mini, and GPT-o4-mini on BN$\rightarrow$CCP and CCP$\rightarrow$BN. 
Results are reported as mean $\pm$ standard deviation for BLEU and chrF.
}
\label{table:icl_zero_shot}
\end{table*}

% \begin{figure*}[ht]
%     \centering
%     \includegraphics[width=\linewidth]{figures/CCP-BN-examples-translation.pdf}
%     \caption{Showing an example of prediction done by BanglaT5 (IBT) on CCP$\rightarrow$BN and a zero-shot translation on EN-CCP trained on BN$\rightarrow$CCP. For CCP$\rightarrow$BN, we mark the wrong words as red, and in the zero-shot translation, we mark the same context words with the same color.}
%     \label{tab:zero-shot_example}
% \end{figure*}

\end{document}